\documentclass[lettersize,journal]{IEEEtran}
\usepackage{amsmath,amsfonts}
\usepackage{algorithmic}
\usepackage{algorithm}
\usepackage{array}
\usepackage[caption=false,font=normalsize,labelfont=sf,textfont=sf]{subfig}
\usepackage{textcomp}
\usepackage{stfloats}
\usepackage{url}
\usepackage{verbatim}
\usepackage{graphicx}
\usepackage{cite}

\usepackage{multirow}
\usepackage{multicol}
\usepackage{booktabs}
\usepackage[section]{placeins}
\usepackage{color}
\usepackage{threeparttable}

\newcommand{\Rmnum}[1]{\expandafter\@slowromancap\romannumeral #1@}
\usepackage{xcolor}

\begin{document}

\title{Iterative Adversarial Attack on Image-guided Story Ending Generation}

\author{Youze Wang, Wenbo Hu, and Richang Hong,~\IEEEmembership{Member,~IEEE,}
\thanks{
Manuscript received 11 July 2023; revised 11 September 2023 and 15
November 2023; accepted 11 December 2023. This work was supported by the National Key Research and Development Program of China (No. 2022ZD0118802), the NSF of China Project (No. 61932009 and 62306098), and the Open Projects Program of State Key Laboratory of Multimodal Artificial Intelligence Systems. (Corresponding author: Wenbo Hu).

Y. Wang, W. Hu, R. Hong are at the School of Computer Science and Information Engineering, Hefei University of Technology, Hefei 230009, China. (E-mail: wangyouze@mail.hfut.edu.cn, \{wenbohu,hongrc\}@hfut.edu.cn) }}


\markboth{IEEE Transactions on Multimedia}%
{Shell \MakeLowercase{\textit{et al.}}: A Sample Article Using IEEEtran.cls for IEEE Journals}


\maketitle

\begin{abstract}
Multimodal learning involves developing models that can integrate information from various sources like images and texts. In this field, multimodal text generation is a crucial aspect that involves processing data from multiple modalities and outputting text. The image-guided story ending generation (IgSEG) is a particularly significant task, targeting on an understanding of complex relationships between text and image data with a complete story text ending. Unfortunately, deep neural networks, which are the backbone of recent IgSEG models, are vulnerable to adversarial samples. Current adversarial attack methods mainly focus on single-modality data and do not analyze adversarial attacks for multimodal text generation tasks that use cross-modal information. To this end, we propose an iterative adversarial attack method (Iterative-attack) that fuses image and text modality attacks, allowing for an attack search for adversarial text and image in a more effective iterative way. Experimental results demonstrate that the proposed method outperforms existing single-modal and non-iterative multimodal attack methods, indicating the potential for improving the adversarial robustness of multimodal text generation models, such as multimodal machine translation, multimodal question answering, etc. 
\end{abstract}


\section{Introduction}
\IEEEPARstart{M}{ultimodal} learning aims to build models that can process and integrate information from multiple modalities, such as image and language, which is an increasing research field with great potential for artificial general intelligence~\cite{baltruvsaitis2018multimodal,radford2021learning}. 
The multimodal text generation task takes data from multiple modalities as input and ends up with text as output, which is considered as the basic ability of human intelligence.
The potential applications of multimodal text generation are far-reaching and transformative, whose innovation and expansion have been into new fields, reshaping the way we communicate and interact with information-rich content.
 Typical applications include multimodal machine translation~\cite{su2019unsupervised, liu2021gumbel}, multimodal dialogue response generation~\cite{sun-etal-2022-multimodal, wang2021modeling}, multimodal question answering~\cite{singh2021mimoqa}, multimodal MemexQA~\cite{47871} and image-guided story ending generation~\cite{huang2021igseg, xue2022mmt}.
Among these tasks, image-guided story ending generation is a natural task for an average person to understand and generate multimodal information and represents a fundamental problem for machine intelligence.
This task introduces ending-related images to the story ending generation,  which can supplement the story ending with diverse visual concepts. IgSEG models requiring understanding complex relationships between text modality data and image modality data is a standard multimodal text generation task.

The deep neural networks, the backbone model of the recent IgSEG models,  albeit rapidly developing, have been shown vulnerable to adversarial samples which include adding imperceptible perturbations on original images~\cite{szegedy2013intriguing, gao2021push} or modifying some words in original texts that do not affect human semantic understanding~\cite{li2020bert, gao2018black}.
Regarding multimodal tasks especially multimodal text generation models, to our knowledge, there has been no research to systematically analyze the adversarial robustness performance and design an adversarial attack solution that utilizes the cross-modal information.
The previous adversarial attack methods mainly focus on the single-modal data, such as image modality adversarial attack methods~\cite{chen2021heu, goodfellow2014explaining}, text modality adversarial attack methods~\cite{li2020bert, li2018textbugger, cheng2020seq2sick}, or simply use a step-wise mechanism which first perturbs the discrete texts and then perturbs the continuous image based on the text perturbation, which is difficult to find the most vulnerable multimodal information patch pairs.

\begin{figure*}[t]
\centering\includegraphics[width=1.7\columnwidth]{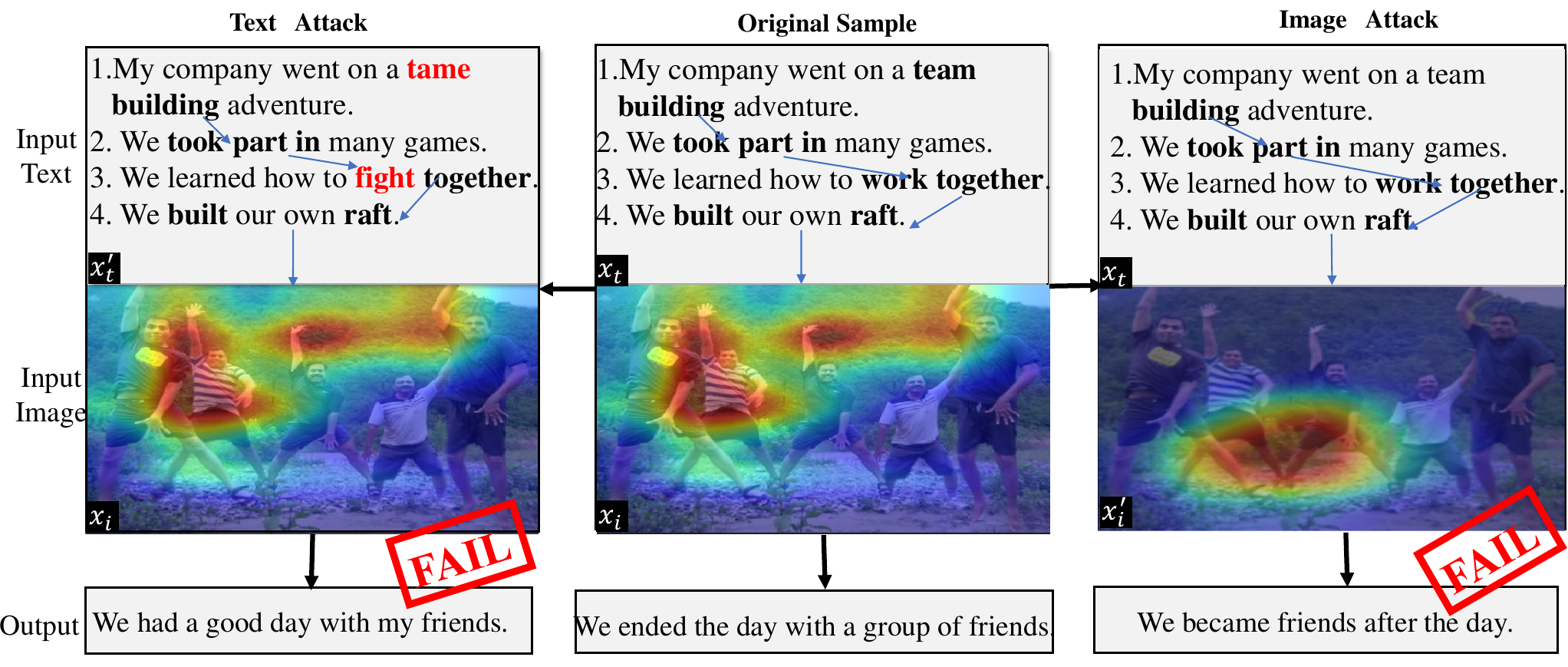}
\caption{An example of the bottleneck single modality adversarial attack against the multimodal text generation model. The blue arrows denote the information chain, and the heat map shows where the target model focuses on the image. When a single modality adversarial example attacks the target model, the other unperturbed modality data may provide complementary information, making the attack fail.}
\label{fig:intro}
\end{figure*}

As we know, the input of multimodal text generation tasks involves inputs from multiple modalities, which is more challenging than the single-modal tasks. 
Simply migrating the single-modal adversarial attack methods may face performance bottlenecks since the information shift caused by the perturbation may be corrected by data for another modality.
For example, Figure~\ref{fig:intro} shows the multimodal information for a story including text context (i.e. story context) and visual information (i.e. ending-related image), for which the single-modal adversarial attack methods all failed due to the complementary information between text data and image data, where the information shift caused by a single-modal adversarial attack can be
corrected by another modality data.
Therefore, the critical issue examined in this paper: 
\emph{how to find the most vulnerable adversarial patch that could take advantage of the cross-modal information?}
To tackle this issue, we need to consider the gap between the discrete text modality and the continuous image modality, it is hard to optimize the designed object function in discrete space. 

In this paper, to investigate the adversarial robustness problem for the image-guided story ending generation task, we propose an iterative adversarial attack method to effectively
craft an imperceptible attack for text-image pair samples. 
We fuse the image modality attack into the text modality attack and iteratively perturb image modality when every small perturbation in the original text modality is changed into continuous space, which allows for an attack search for the adversarial text and image for IgSEG models instead of independently.
Experimental results show that Iterative-attack outperforms the existing single-modal adversarial attack method (kNN and WordSwap ~\cite{michel2019evaluation}) and multimodal adversarial attack method ~\cite{zhang2022towards} in terms of success rate, and semantic similarity. 

In summary, our contribution is as follows:
\begin{itemize}
    \item To the best of our knowledge, this is the first multimodal adversarial attack on a multimodal text generation task.
    \item We propose a new iterative multimodal adversarial attack against IgSEG models to fuse the image modality attack into a text modality attack, which can iteratively find the most vulnerable multimodal information patch.
    \item We evaluate our iterative attack method on four IgSEG models with two real-world datasets, and experimental results demonstrate our approach outperforms the baseline methods.
\end{itemize}

The rest of the paper is organized as follows. Section~\ref{sec:related_work} summarises the related works on the image-guided story ending generation task and adversarial attack including single-modal adversarial attacks and multimodal adversarial attacks. In Section~\ref{sec:method}, we first demonstrate the problem definition in our proposed method. Then we introduce the proposed multimodal adversarial attack algorithm for IgSEG models in detail. In Section~\ref{sec:experiments}, we first conduct a multimodal adversarial attack on IgSEG models with different attack methods. Then we conduct ablation experiments to evaluate the effectiveness of the proposed Iterative-attack. After that, we present some visualization results and discussions to further analyze our method. Section~\ref{sec:conclusion} concludes the paper.
\section{Related Works}
\label{sec:related_work}
\subsection{Image-guided Story Ending Generation Task}
Recently, while multimodal tasks have attracted significant attention~\cite{liu2021hit, wang2020fake}, Huang et al.~\cite{huang2021igseg} proposed a new task called Image-guided Story Ending Generation (IgSEG), which generates a story ending for the story context and an ending-related image.
Huang et al.~\cite{huang2021igseg} first explored incorporating a context and an image to generate a story ending, which proposed a GCN-based text encoder and an LSTM-based decoder. Xue et al.~\cite{xue2022mmt} proposed an end-to-end Multimodal Memory Transformer that modeled and fused both contextual and visual information to obtain the multimodal dependency for IgSEG. 
Other multimodal text generation tasks, such as IgSEG,  also take data from multiple modalities as input and end up with texts as output, which utilize complementary multimodal information. The adversarial attack against multimodal text generation tasks requires finding the most vulnerable multimodal information patch.

\subsection{Adversarial Attack}
Recently, the adversarial attack against Deep Neural Networks (DNN) has drawn the keen interest of researchers~\cite{zhang2021targeted}. The deep neural networks are found vulnerable to adversarial samples, for which the small perturbations are added on the original inputs\cite{cheng2020seq2sick,dong2018boosting, goodfellow2014explaining, madry2017towards,  jin2019bert, ren2019generating, li2020bert, wallace2020imitation, du2021robust}. 
That is, the attack on adversarial samples is imperceptible to human judges while they can mislead incorrect outputs of the deep neural networks. Now, there have been many works done on different data types, such as images, texts, and graphs. Based on the difference in modality, we roughly summarize existing adversarial attack models into two categories: single-modal and multimodal adversarial attacks.

\subsubsection{Single-modal Adversarial Attack}
The adversarial attack is first proposed in computer vision for classification tasks, which illustrates the vulnerability of deep learning models. In the image classification task, there are many algorithms based on both the architecture and the parameters of the model performing gradient-based optimization on the input and constructing adversarial examples, such as  FGSM~\cite{goodfellow2014explaining}, PGD~\cite{madry2017towards},  MIM~\cite{dong2018boosting} and AutoMA~\cite{9599534}. 
In the text classification task, current successful adversarial attack methods adopt heuristic rules to modify the characters of a word~\cite{jin2019bert}, and substitute words with synonyms~\cite{ren2019generating}. 
Gao et al.~\cite{gao2018black} applied perturbations based on word embeddings such as Glove~\cite{pennington2014glove}, which were not strictly semantically and grammatically coordinated.
Li et al.~\cite{li2020bert} turned BERT against its fine-tuned models and used BERT to generate adversarial samples for texts.
In addition, Wang et al.~\cite{wang2022di} explored a white-box adversarial attack for images classifiers from the perspective of interpretable features. Shen et al.~\cite{shen2021bbas} applied boosting-based black-box attacks to enhance the diversity of perturbation, which can contribute to adversarial training. Naseer~\cite{naseer2020self} trained a purifier network in a self-supervised manner to defend against unseen adversarial attacks. During the training, the adversarial images are generated by a self-supervised perturbation that can disrupt the deep perceptual features.

For text generation tasks, universal adversarial attack~\cite{wallace2020imitation}, a new type of attack, consists of a single snippet of text that can be added to any input sentence to mislead the neural machine translation model. Seq2Sick~\cite{cheng2020seq2sick} is a white-box attack method against sequence-to-sequence models, which solves an optimization problem by gradient projection. T3~\cite{wang2020t3} is a tree-based autoencoder to embed the discrete text data into a continuous representation space, which can perform an adversarial perturbation against QA models. CLAPS~\cite{lee2020contrastive} can mitigate the conditional text generation problem by contrasting positive pairs with negative pairs, such that the model is exposed to various valid or incorrect perturbations of the inputs, for improved generalization. 


However, the input of multimodal models involving multiple modalities and the complexity of the text generation tasks make it impractical to directly employ the standard single-modal adversarial attack methods against multimodal text generation tasks.
\subsubsection{Multimodal Adversarial Attack}
Multimodal learning ~\cite{boateng2020towards, long2021improving, huang2021igseg} aims to understand the current scene from multiple modalities. There are some adversarial attacks attempted on multimodal neural networks. Xu et al.~\cite{xu2018fooling} investigated attacking the visual question answering model by perturbing the image modality. Agrawal et al.~\cite{agrawal2018don} and Shah et al.~\cite{shah2019cycle} attempted to attack the vision-and-language model by perturbing the text modality.
Yan et al.~\cite{yang2021defending} showed that standard multimodal fusion models were vulnerable to single-source adversaries, and studied an adversarial robust fusion strategy and proposed a defense method. 
Zhou~\cite{zhou2023advclip} evaluated the vulnerability of CLIP to the universal attack on the image-text retrieval task and the image classification task.
Wang et al.~\cite{wang2023targeted} first explored the targeted adversarial attack against cross-modal hash retrieval, which collaborates with the semantic translator to generate adversarial examples that contain the target semantics specified by the attacker. Zhu et al.~\cite{zhu2023efficient} proposed a query-based multi-modal knockofs-driven adversarial samples generation method to attack cross-modal hash retrieval in a black-box setting.
Zhang et al.~\cite{zhang2022towards} studied the adversarial attack on popular vision-language pre-training models and vision-language tasks. 
However, regarding multimodal text generation tasks, to our knowledge, there is no relevant work to systematically analyze and design adversarial attacks. Compared with the above adversarial attack for multimodal classification tasks, the adversarial attack for multimodal text generation is more challenging.
\section{Method}
\label{sec:method}

\begin{table}[t]
\begin{small}
\centering
  \caption{The main notations of our proposed method.}
\begin{tabular}{c|p{0.8\columnwidth}}
\toprule

Notation&  Description\\ 
\midrule
$\mathcal{F}$& the pre-trained IgSEG models\\ 
$(x_t, x_i)$& the original story context and ending-related image\\
$(x'_t, x'_i)$& the adversarial context and adversarial image\\
 $y$&the ground-truth story ending\\
$y_p$&the story ending generated by the pre-trained IgSEG models with ($x_t, x_i$)\\
 $y_f$&the story ending generated by the pre-trained IgSEG models with ($x'_t, x'_i$)\\
$\mathcal{X}$&  the original multimodal feature space\\
$\mathcal{Y}$& the target text space\\
$n$& the length of the story ending\\
 $w_i$& the $i_{th}$ word in the text\\ 
 $Q_{wh}$& the importance score of $h_{th}$ word in the $x_t$\\
 $Q_x$& the importance score of all words in $x_t$ in descending order\\
$L$&the selected top-K important words from $Q_x$\\
$S_c$&the perturbation in character-level\\
$S_w$&the perturbation in word-level\\
$C_{wh}$& the substitutes set for the import word $w_h$\\
$c_j$&the $j_{th}$ substitute word in $C_{wh}$\\
$\epsilon$& the step size of the image attack\\
$\theta$&  the parameters of the target models\\
$S \in R^d$&   a set of allowed perturbations for image attack\\
$\lambda$&  the threshold for determining whether an attack is successful\\
sign(x)&  a mathematical function that returns
the sign of a real number ”x”\\ 
\midrule
\end{tabular}
\label{tab:notations}
\end{small}
\end{table}

In this section, we detail the proposed iterative multimodal adversarial attack method (Iterative-attack) for IgSEG models.
Our method is inspired by the idea: a multimodal neural network incorporates multimodal information from different modalities that can complement each other. Simply migrating the single-modal adversarial attack methods to multimodal text generation tasks would face the dilemma that the information shift caused by a single-modal adversarial attack may be corrected by another modality's information. 
To solve the problem, Iterative-attack fuses the image modality attack into the text modality attack to iteratively find the most vulnerable multimodal information patch.
The key notations used in the paper as summarized in Table \ref{tab:notations}.

\subsection{Problem Formulation}
Given a pre-trained IgSEG model $\mathcal{F} : \mathcal{X} \rightarrow \mathcal{Y}$, which maps from the origin multimodal feature space $\mathcal{X}$ to the target text space $\mathcal{Y}$. The IgSEG model $\mathcal{F}$ generally has an DNN-based encoder-decoder structure~\cite{huang2021igseg, xue2022mmt} and aims to maximize the story ending generation probability $p(y|x)$, where $x \in \mathcal{X}$ is the input story context $x_t$ and ending-related image $x_i$ in the origin multimodal space, and $y \in \mathcal{Y}$ is the ground-truth story ending in target text space, where $\mathcal{F}((x_t, x_i)) = y_p$.
A successful multimodal adversarial attack against IgSEG models is to generate an adversarial context $x'_t$ and an adversarial image $x'_i$, so that the BLEU score of the story ending generated by taking the adversarial sample ($x'_t, x'_i$) as input relative to the BLUE score of the original story ending is less than a threshold $\lambda$.

\subsection{Multimodal Adversarial Attack Loss}
By perturbing texts or images, generating adversarial examples against DNN models can fool DNN models. However, single-modal  adversarial attack can't effectively maximize the attack on the output of multimodal models~\cite{zhang2022towards}. We address this issue by developing an effective iterative multimodal adversarial attack method against IgSEG models.

To find a more effective adversarial input $(x'_t, x'_i)$, we try to maximize the adversarial loss of the target IgSEG model. Since the IgSEG models are trained to generate the next token of the story ending given the ending up until that token, we are looking for the adversarial text and image that can maximize the probability of wrong story ending (i.e., minimizes the probability of correct story ending) for the $i$-th token, given that the IgSEG model has produced the correct story ending up to step ($i$-1). We can calculate the adversarial loss as the following loss function:
\begin{equation}
    \mathcal{L}_{adv} = \frac{1}{n}\sum_{i=1} ^{n} \log p\left(y_i | (x'_t, x'_i), {y_1, ...,y_{(i-1)}}\right),
    \label{eq:adversarial_loss}
\end{equation}
where $n$ is the length of the story ending.
By minimizing $\log p(\cdot)$, normalized by the sentence length $n$, we force the output probability vector of the IgSEG model to differ from the delta distribution on the token $y_i$, which may cause the predicted story ending to be wrong.


\begin{figure}[t]
\centering\includegraphics[width=3in]{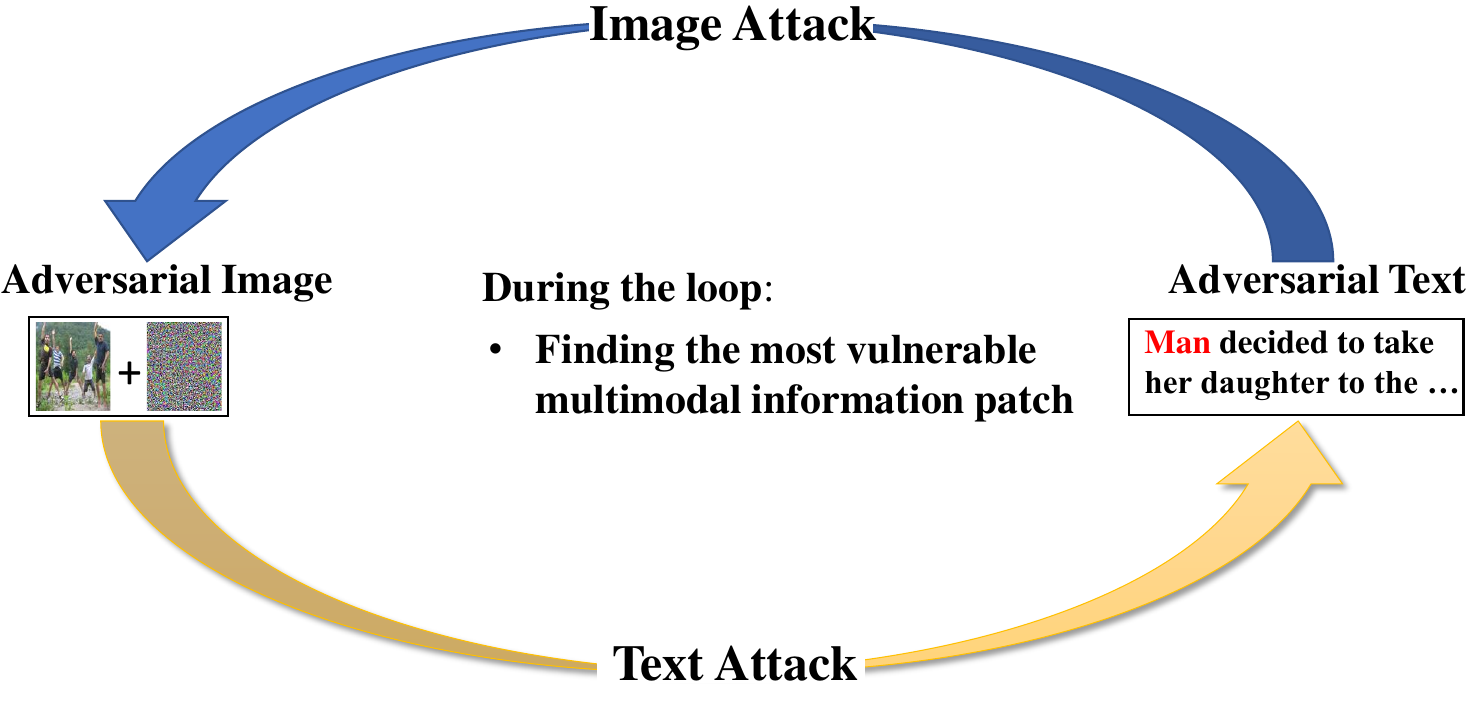}
\caption{Illustration of Iterative-attack. We fuse the image modality attack into the text modality attack to iteratively find the most vulnerable multimodal information patch, which can avoid the dilemma that the information shift caused by a single-modal adversarial attack may be corrected by another modality’s information.}
 \label{fig:method}
\end{figure}
\subsection{Adversarial Attack Algorithms}
We now introduce the proposed algorithm that generates adversarial examples including adversarial text and adversarial image for IgSEG models. We first need to generate potential  adversarial texts as supervised information, and then iteratively take the potential adversarial texts as input to attack the relevant image information to find the most vulnerable multimodal information patch. The flowchart of Iterative-attack is shown in Algorithm 1.

\textbf{Step 1: Generate Potential Adversarial Text.}  
Potential adversarial texts can be used as the input of Iterative-attack to find complementary adversarial images. In general, if a word is more important to the output of the target model,  it is more likely to be attacked. Motivated by~\cite{li2020bert}, we select the most important words in the story context which have a higher significant influence on the final output logit. 

Specifically, Let $x_t = [w_0, ..., w_h, ...]$ , and $\mathcal{F}((x_t, x_i)$ denotes the normal output of the IgSEG model, the importance score $Q_{w_h}$ of word $w_h$ in the original text $x_t$ is defined as:
\begin{equation}
    Q_{w_h} = \mathcal{F}((x_t, x_i)) - \mathcal{F}((x_{t_{@h}}, x_i))
\label{eq:findingImportantWords}
\end{equation}
where $x_{t_{@h}} = [w_0, \cdots, w_{h-1}, \text{MASK}, w_{h+1} ...]$ is the story context after replacing $w_h$ with a special character [MASK]. We can obtain the important score of every word for the target model by using the Equation.~\ref{eq:findingImportantWords}.
The important score of all words in the original text is $Q_x$, which is defined as (Algorithm 1, line 2 - 4) :
\begin{equation}
    Q_x = [Q_{w_1}, \cdots, Q_{w_h}, \cdots, Q_{w_n}]
\end{equation}
Then, we sort $Q_x$ in descending order and select Top-K words as the important word list $L$ based on the descending order of $Q_x$ (Algorithm 1, line 5).

To generate candidate perturbation for every important word in list $L$ and ensure that the generated adversarial text is visually or semantically similar to the benign one for human understanding,
we combine bugs generation ~\cite{li2018textbugger} in character-level perturbation called $S_c$ and words replacement via BERT ~\cite{li2020bert} in word-level perturbation called $S_w$ as the text perturbation mechanism of Iterative-attack. 
The substitutes for the important word $w_h$ is $C_{w_h} = S_c \oplus \footnote{means concatenation } S_w$ , which will be used to generate a potential adversarial context as the input of iterative multimodal adversarial attack in step 2.

\textbf{Step 2: Iterative Multimodal Adversarial Attack}.
To effectively attack the IgSEG models, we propose a new iterative multimodal adversarial attack solution, which fuses the image attack into the text attack, as shown in Figure~\ref{fig:method}. 
We iteratively attack the image and text inputs $(x_{t},x_{i})$ to find the most vulnerable multimodal information patch until the story ending generation quality of the adversarial example ($x'_t, x'_i$) relative to the original story ending generation quality is less than a threshold $\lambda$.

For the text adversarial attack, we generate substitute words set $C_{w_h}$ for the vulnerable word $w_h$ obtained in Step 1 (Algorithm 1, line 7). For every substitute word $c_j$ in $C_{w_h}$, we replace $w_h$ in the original context with  $c_j$ to generate the potential adversarial text $x'_t$ (Algorithm 1, line 10 - 11). 

For the image  adversarial attack,  we try to perturb image information that is complementary to the perturbations in the text, and output the potential adversarial image as follows (Algorithm 1, line 12):
 \begin{equation}
    x_i^{' (a+1)} = \Pi_{x_i + S}(x_i^{'a} + \epsilon \cdot \mathrm{sign}(\bigtriangledown _{x_i} \mathcal{L}_{adv}(\theta, (x'_t, x_i), y)))
\end{equation}
where $a$ is the steps of iteration, $\epsilon$ is the step size, $\theta$ is the parameter of target models, $S\subseteq R^d$ is a set of allowed perturbations, $\mathrm{sign}$(x)  is a mathematical function that returns the sign of a real number "x". Based on potential adversarial text $x'_t$, perturb the image $x_i$ to find the most vulnerable multimodal information pairs ($x'_t, x'_i$) on the target model, thereby affecting the model's output that significantly differs from the ground truth label $y$. 

When the BLEU score of the story ending generated by taking the adversarial sample ($x'_t, x'_i$) as input relative to the BLEU score of the original story ending is less than the threshold $\lambda$, which shows the multimodal adversarial attack on target models is successful as follows:
\begin{equation}
    \frac{BLEU(\mathcal{F}((x_t^{'}, x_i{'})), y)}{BLEU(\mathcal{F}((x_t, x_i)), y)} \leq \lambda .
    \label {eq:BLEU_evalution}
\end{equation}
where $\mathbf{}{BLEU}$ is a function that calculates the BLEU value~\cite{papineni2002bleu} between the story ending outputted by the target model $\mathcal{F}$ and the ground truth label $y$.  Equation~\ref{eq:BLEU_evalution} reflects the relative percentage decrease in IgSEG model performance against multimodal adversarial samples. 

Otherwise, saving the loss between $y_f$ and $y$, and the potential adversarial text $x'_t$. We continue to attack the potential adversarial text which replaces the word $w_h$ with the next substitute word $c_{j+1}$ (Algorithm 1, line10 - 19). 

If all the substitute words in $C_{wh}$ can not satisfy the Eq.~\ref{eq:BLEU_evalution}, we will continue to iteratively attack the next important word $w_{h+1}$ based on the text where the word $w_h$ is replaced by the substitute word in $C_{w_h}$ maximizing the adversarial loss (Algorithm 1, line 20 - 23).

\section{Experiments}
\label{sec:experiments}

In this section, we evaluate our proposed Iterative-attack method by applying it to four pre-trained IgSEG models on two datasets: VIST-E~\cite{huang2021igseg} and LSMDC-E~\cite{xue2022mmt}. For comparative analysis, three baseline methods are also employed. To assess the individual components' contribution within our method, we have conducted an ablation study. Additionally, we examine the influence of hyper-parameters on the Iterative-attack's performance by varying the number of word perturbations and the count of important words, analyzing their impact on runtime. To further analyze the effectiveness of Iterative-attack, we test our method in a multimodal machine translation dataset.
Finally, we present two visualizations of multimodal adversarial samples, accompanied by an error analysis, to offer an intuitive understanding of the Iterative-attack approach. 

\subsection{Experimental Setup}

\textbf{\emph{Dataset}}. This work utilizes the VIST-E~\cite{huang2021igseg} and LSMDC-E~\cite{xue2022mmt} datasets to evaluate the task of generating image-guided story endings. The VIST-E dataset is derived from the VIST dataset and consists of 49,913 training samples, 4,963 validation samples, and 5,030 test samples. Unlike the original VIST dataset, each sample in VIST-E specifically contains the story ending, the corresponding image related to the ending, and the first four sentences that constitute the story context. To ensure uniformity, the length of each sentence is limited to a maximum of 40 words.
Similarly, the LSMDC-E dataset is derived from the LSMDC 2021 dataset~\cite{rohrbach2015dataset} and comprises 20,151 training samples, 1,477 validation samples, and 2,005 test samples. In LSMDC-E, the story context is constructed by selecting the first four sentences from each five-sentence story, while the last sentence is designated as the story ending. As each sentence in LSMDC-E is associated with a set of movie frames, the set corresponding to the last frame is chosen as the ending-related image set. To maintain consistency with prior work~\cite{huang2021igseg}, a maximum sentence length of 20 words is imposed. Due to the LSMDC-E dataset is not publicly available, we constructed the dataset as described in~\cite{xue2022mmt}.
Both of the VIST-E and LSMDC-E datasets are widely recognized benchmarks for evaluating the task of Image-guided story ending generation. They serve as effective measures for assessing a model's capacity to comprehend multimodal information and generate text.

The Multi30k~\cite{elliott2016multi30k}  dataset, an expansion of the original Flickr30k~\cite{plummer2015flickr30k}, is a pivotal dataset in multimodal machine translation research. It comprises 30,000 images, each accompanied by textual descriptions in both English and German. The dataset consists of two primary variants: M30kT and M30KC. M30kT features each image with a single English description professionally translated into German. In contrast, M30KC provides five English and five German descriptions for each image, with the German texts sourced independently via crowdsourcing, rather than direct translations. This dataset is partitioned into training, validation, and test sets containing 29,000, 1,014, and 1,000 instances, respectively.  In the attack, we present experiment results on the English-German (En-De) Test2016.

\begin{algorithm}[ht]
	\renewcommand{\algorithmicrequire}{\textbf{Input:}}
	\renewcommand{\algorithmicensure}{\textbf{Output:}}
	\caption{Iterative-attack Algorithm}
	\label{alg}
	\begin{algorithmic}[1]
		\REQUIRE 
        Original sample $X = \{x_t, x_i\}$; Ground-truth ending of a story $Y$; Normal story ending $Y_p$ = $\mathcal{F}((x_t, x_i))$; The maximum number of iterations for generating adversarial visual sample $N_{iter}$;  The number of perturbing words $P$; The number of important words K;  The threshold $\lambda$;
		\ENSURE  A multimodal adversarial sample $X' = \{x'_t, x'_i\}$;
		\STATE Initialize: image\_{attacker} : iterative multimodal adversarial attack; 
        \FOR {word $w_a$ in $x_t$}
        \STATE  Calculate importance score $Q_{x_a}$ according to Eq.~\ref{eq:findingImportantWords}
        \ENDFOR
        \STATE create important word list $L$ $\leftarrow$ $[w_{top-1}, w_{top-2}, \cdot \cdot \cdot ,w_{top-K}]$ according to $Q_x$
        \FOR{ $w_h$ in $L$} 
        \STATE Generate substitutes set $C_{w_h}$ for word $w_h$: $C_{w_h} = S_c \oplus S_w$
        \STATE $U \leftarrow$ empty set for saving the loss between $Y_f$ and $Y$; 
        \STATE $V \leftarrow$ empty set for saving the adversarial story context;
        \FOR{ $c_j$ in $C_{w_h}$}
        \STATE $x'_t$ $\leftarrow$ replace word $w_h$ with $c_j$ 
        \STATE $x'_i$ $\leftarrow$ image\_{attacker} $(x'_t, x_i, N_{iter})$
        \STATE Generate story ending with adversarial sample $(x'_t, x'_i)$: $Y_f = \mathcal{F} ((x'_t, x'_i))$
        \IF{ $\frac{BLEU((Y_f, Y))}{BLEU(Y_p, Y)}\leq \lambda $}
        \STATE return $(x'_t, x'_i)$
        \ELSE
        \STATE add $\mathcal{L} (Y_f, Y)$ to set $U$; add $x'_t$ to set $V$;
        \ENDIF
        \ENDFOR
        \STATE $x_t = V[t]$, where $t$ = argmax$(U)$
        \STATE h += 1
        \IF{$h > P$}
        \STATE return None
        \ENDIF
        \ENDFOR
        \RETURN None
    \end{algorithmic}
\end{algorithm}

\textbf{\emph{Hyper-parameters}}. 
For the perturbation on images, we apply the PGD attack~\cite{madry2017towards}. The maximum perturbation $\epsilon$ is set to 4/255, and the step size is set to $\epsilon$/10. The number of PGD iterations is set to 20. For a fair comparison, the maximum number $P$ of the perturbed words in the text is set to 2 for all adversarial attack methods. The number of important words K is set to 10; Following the setting in~\cite{sadrizadeh2023transfool}, the threshold $\lambda$ is set to 0.5. In the multimodal machine translation task, the maximum number $P$ of the perturbed words in the text is set to 1 for all adversarial attack methods due to the texts in Multi30k dataset is shorter than texts in VIST-E and LSMDC-E datasets. 
\subsection{IgSEG Methods and Attacking Baselines}
\textbf{\emph{IgSEG Methods}}.
To prove the effectiveness of the proposed multimodal adversarial attack method for IgSEG models, we select Seq2Seq~\cite{luong2015effective}, Transformer~\cite{vaswani2017attention}, MGCL~\cite{huang2021igseg}, and MMT~\cite{xue2022mmt} as the target models.
On the VIST-E dataset and LSMDC-E dataset, we 
reimplement four IgSEG methods based on the official open source codes or settings in the original papers, where we apply four widely-used automatic: BLEU~\cite{papineni2002bleu}, METEOR~\cite{banerjee2005meteor}, CIDEr~\cite{vedantam2015cider}, and ROUGE-L~\cite{lin2004rouge} to evaluate the three IgSEG models. The reproducible results are reported in Table~\ref{tab:original_IgSEG_models}. Compared the actual results of running with the report results in the papers, we can observe that the results of the four IgSEG models on two datasets are close to the report results in the paper~\cite{huang2021igseg, xue2022mmt}, which demonstrates the IgSEG models we attack are normal. The four IgSEG models as follows:
\begin{itemize}
    \item \textbf{Seq2Seq}~\cite{luong2015effective} is an attention-based model with stacked RNNs, which utilizes two attentional mechanisms for neural machine translation, where the global attention always looks at all source positions and the local one only attends to a subset of source positions at a time.  To adapt to the IgSEG task, we concatenate textual and visual features as its inputs.
    \item \textbf{Transformer}~\cite{vaswani2017attention} is a parallel model based solely on the attention mechanism and is widely used in text generation tasks.
To adapt to the IgSEG task, we concatenate textual and visual features as its inputs.
\item  \textbf{MGCL}~\cite{huang2021igseg} is the first method to introduce an ending-related image to explore a more informative and reasonable ending, which proposes a GCN-based text encoder and an LSTM-based decoder to build logically consistent and semantically rich story endings. 
\item \textbf{MMT}~\cite{xue2022mmt} extracts the multimodal semantic dependency for IgSEG with a multimodal transformer that can build and fuse visual and contextual information. Besides, a cross-modal attention network is used to learn cross-modal relations and fuse the fine-grained feature.
\end{itemize}

\textbf{\emph{Adversarial Attack Methods}}.
We compare our attack with Co-attack~\cite{zhang2022towards}, which is a multimodal adversarial attack against vision-language pre-training models for classification tasks and image-text retrieval tasks in a non-iterative way; We also adapt the kNN and CharSwap in~\cite{michel2019evaluation}, a white-box untargeted attack against neural machine translation models. kNN substitutes some words with their neighbors in the embedding space; CharSwap considers swapping the characters in the target word.

\textbf{\emph{Variants of Iterative-attack.}}
The following variants of the proposed Iterative-attack are designed for comparison in the ablation experiment.
\begin{itemize}
    \item Text-attack: the variant of Iterative-attack, which removes image adversarial attack when attacking the target models.
    \item Image-attack: the variant of Iterative-attack, which removes text adversarial attack when attacking the target models.
    \item Character-attack: the variant of Iterative-attack, which removes the word-level substitutes generation strategy for vulnerable words when attacking the target models.
     \item Word-attack: the variant of Iterative-attack, which removes the character-level substitutes generation strategy for vulnerable words when attacking the target models.
\end{itemize}

\subsection{Evaluation Metrics}

For evaluation, we adopt a multifaceted approach to performance metrics, similar to~\cite{sadrizadeh2023transfool}. We assess:
(1)\textbf{Attack Success Rate (ASR)}: This metric quantifies the proportion of successful adversarial examples, defined in line with~\cite{ebrahimi2018adversarial} as those with a BLEU score for the story ending less than half that of the original.
\textbf{(2) Relative Decrease in Story Ending Generation Quality (RDBLEU and RDchrF)}: We measure the degradation in story ending generation quality using BLEU score and chrF, computed respectively as RDBLEU and RDchrF~\cite{popovic2015chrf}. We choose to compute the relative decrease in story ending generation quality so that scores are comparable across different models and datasets.
\textbf{(3) Semantic Similarity (Sim.)}: Calculated between original and adversarial sentences using the universal sentence encoder~\cite{yang2020multilingual}, this metric approximates the semantic alteration introduced by the attack.
\textbf{(4) Perplexity Score (Perp.)}: The fluency of adversarial examples is evaluated using the GPT-2 (large) perplexity score.
The whole method is implemented in Pytorch~\cite{paszke2019pytorch}, with all experiments conducted on a GeForce RTX 1080Ti GPU.

\begin{table*}[ht]
\caption{Performance of adversarial attack against different IgSEG models on VIST-E dataset.}
\vspace{2mm}
\centering
\begin{tabular}{c|c|c|ccccc}
\hline
Dataset                  & Method                     & Attack                    & ASR(\%) $\uparrow$        & RDBLEU$\uparrow$         & RDchrF$\uparrow$         & Sim.$\uparrow$           & Perp.$\downarrow$            \\ \hline
\multirow{16}{*}{VIST-E} & \multirow{4}{*}{Seq2Seq}   & \textbf{Iterative-attack} & \textbf{57.09} & \textbf{0.46} & \textbf{0.27} & \textbf{0.96} & \textbf{122.07} \\
                         &                            & Co-attack                 & 25.84          & 0.20          & 0.26          & 0.94          & 149.50          \\
                         &                            & kNN                       & 23.25          & 0.19          & 0.18          & 0.94          & 122.28          \\
                         &                            & CharSwap                  & 31.69          & 0.30          & 0.24          & 0.94          & 171.37          \\ \cline{2-8} 
                         & \multirow{4}{*}{Transformer} & \textbf{Iterative-attack} & \textbf{35.04} & \textbf{0.18} & \textbf{0.06} & 0.93          & 104.43          \\
                         &                            & Co-attack                 & 22.80          & 0.12          & 0.05          & \textbf{0.95} & 159.28          \\
                         &                            & kNN                       & 14.62          & 0.11          & 0.02          & 0.94          & \textbf{100.82} \\
                         &                            & CharSwap                  & 12.63          & 0.08          & 0.02          & 0.91          & 180.56          \\ \cline{2-8} 
                         & \multirow{4}{*}{MGCL}      & \textbf{Iterative-attack} & \textbf{50.37} & \textbf{0.49} & \textbf{0.23} & \textbf{0.96} & 82.95           \\
                         &                            & Co-attack                 & 39.17          & 0.38          & 0.15          & 0.93          & 118.94          \\
                         &                            & kNN                       & 15.45          & 0.19          & 0.14          & 0.93          & \textbf{82.83}  \\
                         &                            & CharSwap                  & 15.30          & 0.11          & 0.12          & 0.88          & 109.23          \\ \cline{2-8} 
                         & \multirow{4}{*}{MMT}       & \textbf{Iterative-attack} & \textbf{39.67} & \textbf{0.30} & \textbf{0.21} & \textbf{0.95} & 82.54           \\
                         &                            & Co-attack                 & 30.01          & 0.25         & 0.18          & 0.94          & 90.28           \\
                         &                            & kNN                       & 14.89          & 0.12          & 0.08          & 0.93          & \textbf{82.22}  \\
                         &                            & CharSwap                  & 15.56          & 0.12          & 0.12          & 0.90          & 102.24          \\ \hline
\end{tabular}
\label {tab:attack_vist}
\end{table*}

\begin{table*}[ht]
\caption{Performance of adversarial attack against different IgSEG models on LSMDC-E dataset.}
\vspace{2mm}
\centering
\begin{tabular}{c|c|c|ccccc}
\hline
Dataset                   & Method                     & Attack                    & ASR(\%)$\uparrow$        & RDBLEU$\uparrow$        & RDchrF$\uparrow$        & Sim.$\uparrow$          & Perp. $\downarrow$          \\ \hline
\multirow{16}{*}{LSMDC-E} & \multirow{4}{*}{Seq2Seq}   & \textbf{Iterative-attack} & \textbf{57.14} & \textbf{0.53} & \textbf{0.29} & \textbf{0.96} & \textbf{126.77} \\
                          &                            & Co-attack                 & 23.72          & 0.20          & 0.15          & 0.94          & 179.82          \\
                          &                            & kNN                       & 3.40           & 0.03          & 0.03          & 0.96          & 243.42          \\
                          &                            & CharSwap                  & 12.04          & 0.09          & 0.07          & 0.93          & 303.33          \\ \cline{2-8} 
                          & \multirow{4}{*}{Transformer}& \textbf{Iterative-attack} & \textbf{31.72} & \textbf{0.28} & \textbf{0.20} & \textbf{0.96} & \textbf{176.30} \\
                          &                            & Co-attack                 & 20.75          & 0.12          & 0.06          & 0.95          & 189.45          \\
                          &                            & kNN                       & 7.02           & 0.01          & 0.01          & 0.93          & 190.42          \\
                          &                            & CharSwap                  & 6.04           & 0.01          & 0.01          & 0.93          & 205.33          \\ \cline{2-8} 
                          & \multirow{4}{*}{MGCL}      & \textbf{Iterative-attack} & \textbf{47.18} & \textbf{0.42} & \textbf{0.22} & \textbf{0.96} & 176.31  \\
                          &                            & Co-attack                 & 25.76          & 0.20          & 0.18          & 0.93          & \textbf{126.98}          \\
                          &                            & kNN                       & 21.08          & 0.22          & 0.12          & 0.75          & 167.69          \\
                          &                            & CharSwap                  & 19.76          & 0.19          & 0.08          & 0.85          & 213.72          \\ \cline{2-8} 
                          & \multirow{4}{*}{MMT}       & \textbf{Iterative-attack} & \textbf{52.34} & \textbf{0.44} & \textbf{0.21} & \textbf{0.96} & 151.75          \\
                          &                            & Co-attack                 & 36.92          & 0.33          & 0.21          & 0.94          & 199.73          \\
                          &                            & kNN                       & 19.04          & 0.17          & 0.10          & 0.90          & \textbf{142.08} \\
                          &                            & CharSwap                  & 20.24          & 0.21          & 0.12          & 0.87          & 187.99          \\ \hline
\end{tabular}
\label {tab:attack_lsmdc}
\end{table*}

\begin{table*}[ht]
\centering
\caption{The performance of the IgSEG models on the VIST-E dataset and LSMDC-E dataset.}
\vspace{2mm}
\begin{tabular}{c|c|ccccccc}
\hline
Dataset                  & Method        & B1    & B2   & B3   & B4   & M     & C     & R-L   \\ \hline
\multirow{8}{*}{VIST-E}  & Seq2Seq\#     & 13.96 & 5.57 & 2.94 & 1.69 & 4.54  & 12.04 & 16.84 \\
                         & Seq2Seq*      & 14.35 & 6.11 & 3.89 & 1.45 & 8.15  & 10.01 & 11.95 \\
                         & Transformer\# & 17.18 & 6.29 & 3.07 & 2.01 & 6.91  & 12.75 & 18.23 \\
                         & transformer*  & 18.26 & 5.76 & 4.02 & 1.69 & 11.80 & 12.31 & 13.44 \\
                         & MGCL\#        & 22.57 & 8.16 & 4.23 & 2.49 & 7.84  & 21.46 & 21.66 \\
                         & MGCL*         & 22.36 & 7.94 & 5.55 & 2.33 & 14.30 & 18.96 & 19.32 \\
                         & MMT\#         & 22.87 & 8.68 & 4.38 & 2.61 & 15.55 & 25.41 & 23.61 \\
                         & MMT*          & 22.65 & 8.64 & 4.41 & 2.53 & 14.93 & 23.17 & 22.12 \\ \hline
\multirow{8}{*}{LSMDC-E} & Seq2Seq\#     & 14.21 & 4.56 & 1.70 & 0.70 & 11.01 & 8.69  & 19.69 \\
                         & Seq2Seq*      & 13.53 & 3.44 & 1.49 & 0.50 & 8.83  & 5.49  & 16.51 \\
                         & Transformer\# & 15.35 & 4.49 & 1.82 & 0.76 & 11.43 & 9.32  & 19.16 \\
                         & transformer*  & 14.11 & 3.71 & 2.21 & 0.65 & 8.88  & 7.09  & 18.94 \\
                         & MGCL\#        & 15.89 & 4.76 & 1.57 & 0.00 & 11.61 & 9.16  & 20.30 \\
                         & MGCL*         & 14.60 & 3.75 & 1.61 & 0.00 & 9.20  & 6.79  & 17.75 \\
                         & MMT\#         & 18.52 & 5.99 & 2.51 & 1.13 & 12.87 & 12.41 & 20.99 \\
                         & MMT*          & 16.85 & 5.58 & 2.10 & 0.96 & 11.07 & 13.05 & 18.75 \\ \hline
\end{tabular}
\centering
\begin{threeparttable}
* indicates the actual results of running. \# indicates the results in the paper.
\end{threeparttable}
\label{tab:original_IgSEG_models}
\end{table*}

\subsection{Quantitative Results}
Table~\ref{tab:attack_vist} and Table~\ref{tab:attack_lsmdc} shows the experimental results~\footnote{We discard the original samples whose BLEU score of generated story ending is zero to prevent improving the results artificially.} of automatic metrics of attacking different IgSEG methods on VIST-E and LSMDC-E datasets. From the Table~\ref{tab:attack_vist} and Table~\ref{tab:attack_lsmdc} we can draw the following main observations:
\begin{itemize}
    \item [(1)]  Overall, The proposed Iterative-attack can decrease the BLEU score of the target model to more than 30\% of the BLEU score of the original story ending for more than 39\% of the stories for MMT, MGCL, and Seq2Seq on two datasets (except for the Transformer model, where the ASR is more than 30\% on two datasets). Also, in all cases, semantic similarity is more than 0.95, which shows that the Iterative-attack can maintain a high level of semantic similarity with the original context.
    
    \item [(2)] When compared to baselines like kNN and CharSwap, Iertaive-attack exhibits a superior attack success rate across different IgSEG architectures, more significantly degrading the quality of generated story endind. The lower performance of kNN and CharSwap in terms of ASR, RDBLEU, and RDchrF suggests that text-only adversarial attacks are insufficient for disrupting multimodal text generation models, highlighting the importance of attacking both modalities.
    \item [(3)] Co-attack, while outperforming single-modal attack methods in multimodal text generation tasks, falls short when compared to the Iterative-attack. The reason can be attributed to its iterative process that integrates image modality attacks into text modality attacks, enhancing its disruptive capability.
    \item [(4)] Multimodal attack methods (Iterative-attack and Co-attack) demonstrate superior performance over single-modal methods (kNN and CharSwap). This confirms that for multimodal text generation models, reliance on a single modality for perturbation is often ineffective due to the models' ability to compensate with complementary information from the other modality. 
    \item [(5)] The adversarial robustness of MGCL has worse performance than that of MMT on VIST-E dataset, where the reason may be that the LSTM-based decoder in MGCL is more sensitive than the transformer decoder in MMT.
    \item [(6)] In summary, these results collectively indicate that Iterative-attack effectively exploits the vulnerabilities of IgSEG models, significantly degrading the quality of generated text while preserving semantic similarity of adversarial texts. The comparative analysis of attack methods highlights the necessity of targeting both text and image modalities to effectively compromise multimodal text generation  systems.

\end{itemize}

 \begin{figure*}[htb]
\centering\includegraphics[width=6.5in]{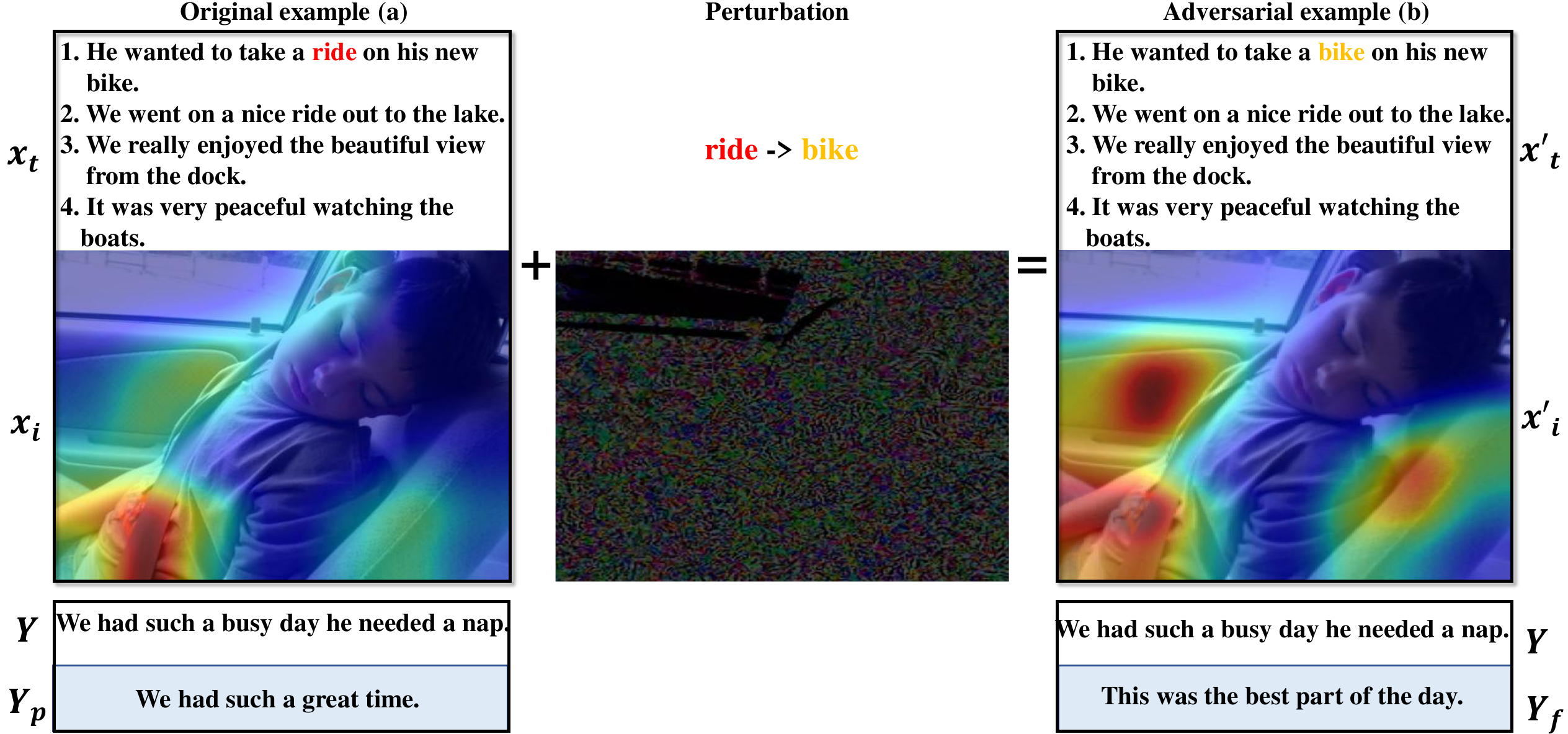}
\caption{The Grad-CAM visualizations of (a) the original example $(x_t, x_i)$, (b) the adversarial example $(x'_t, x'_i)$ derived by Iterative-attack against MMT on VIST-E dataset where the adversarial perturbation is obtained by $x'_i - x_i$ ( pixel values of perturbation are amplified ×20 for visualization).}
 \label{fig:visualization_MMT_1}
\end{figure*}

\begin{figure*}[htb]
\centering\includegraphics[width=6.5in]{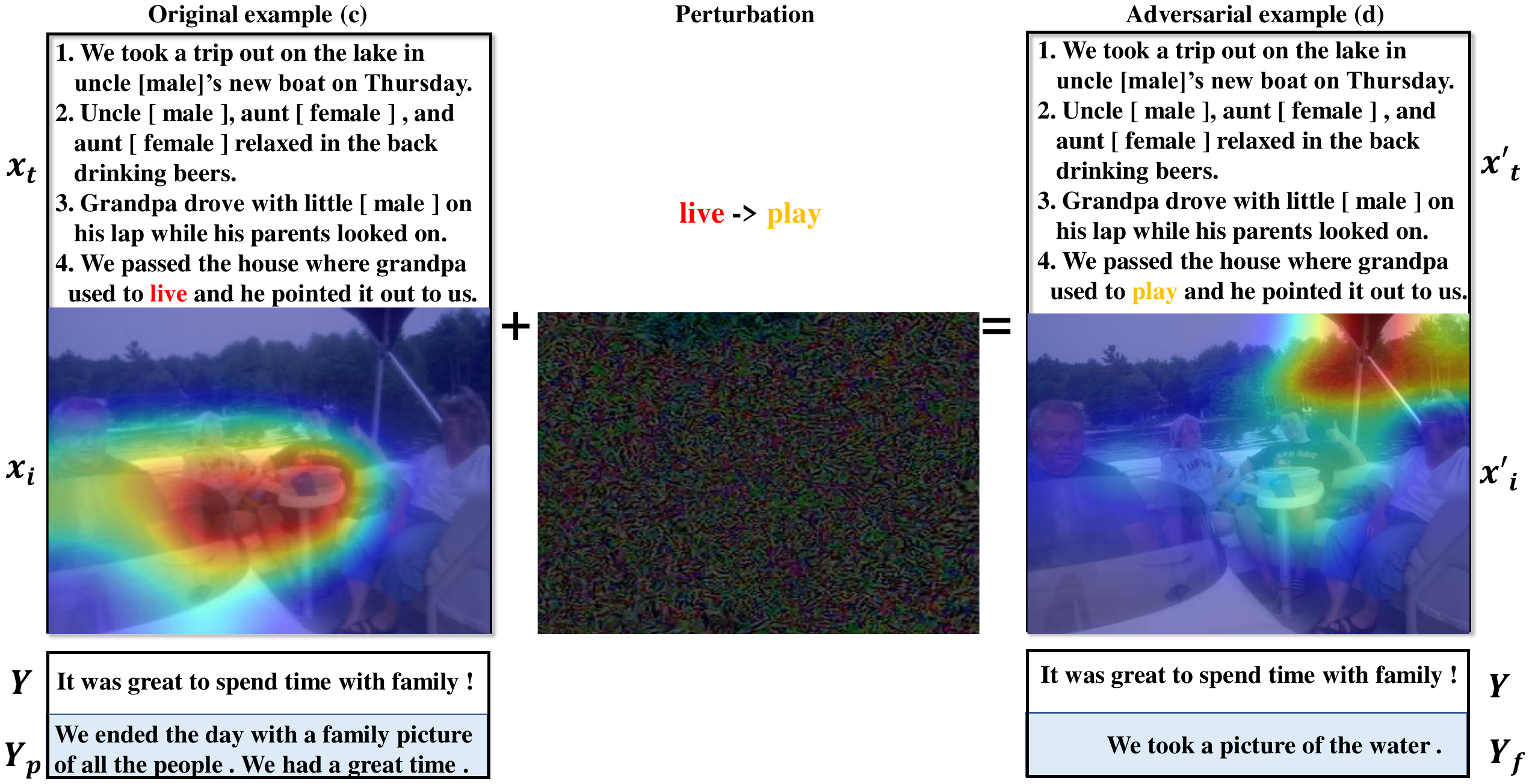}
\caption{The Grad-CAM visualizations of (c) the original example $(x_t, x_i)$, (d) the adversarial example $(x'_t, x'_i)$ derived by Iterative-attack against MMT on VIST-E dataset, where the adversarial perturbation is obtained by $x'_i - x_i$ ( pixel values of perturbation are amplified ×20 for visualization).}
 \label{fig:visualization_MMT_2}
 \end{figure*}

\begin{figure*}[htb]
\centering \includegraphics[width=6.5in]{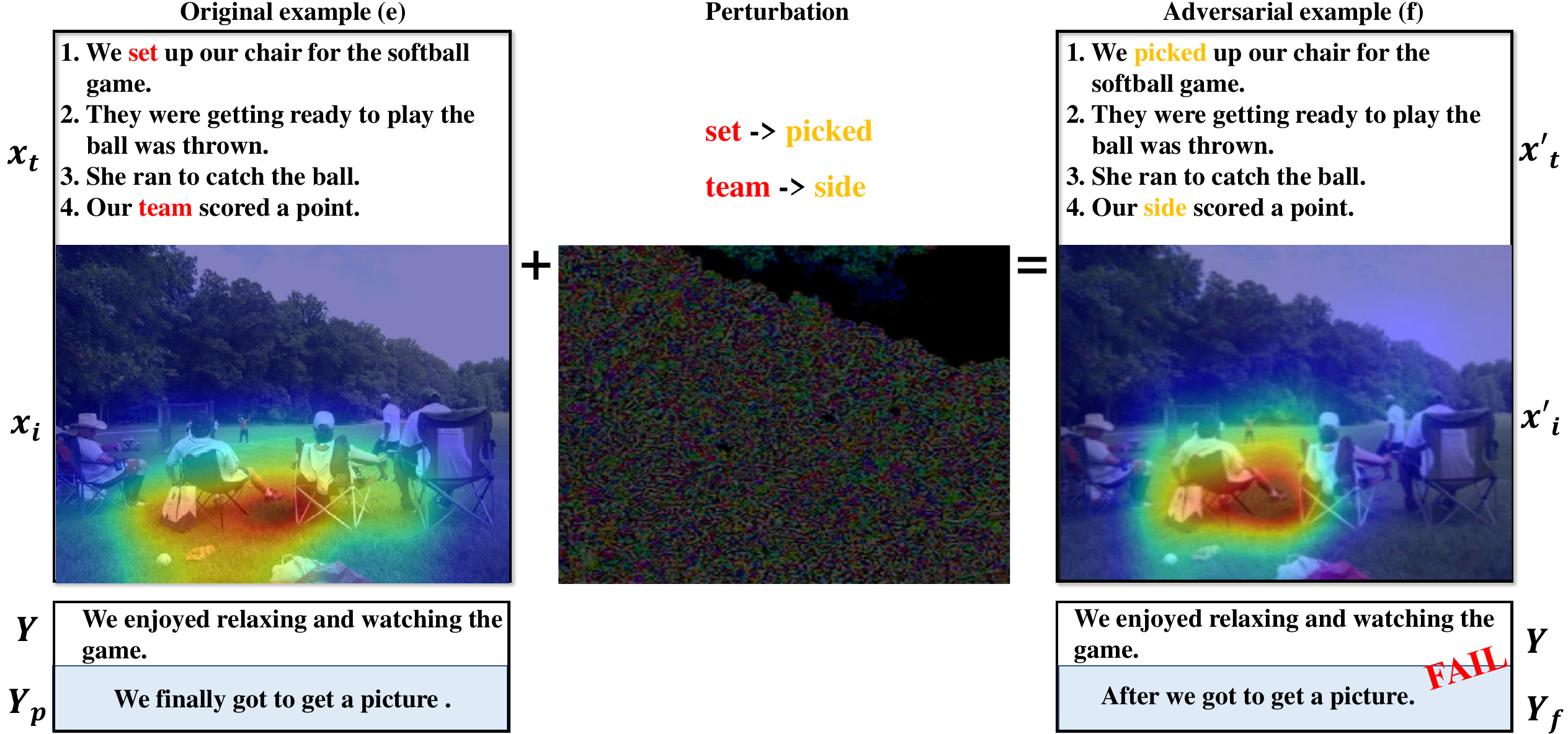}
\caption{The Grad-CAM visualizations of (e) the original example $(x_t, x_i)$, (f) the adversarial example $(x'_t, x'_i)$ derived by Iterative-attack against MMT on VIST-E dataset where the adversarial perturbation is obtained by $x'_i - x_i$ ( pixel values of perturbation are amplified ×20 for visualization).}
 \label{fig:visualization_bad_case}
\end{figure*}

\begin{table*}[ht]
\caption{Results of comparison with different modality attacks in our method against MGCL on VIST-E and LSMDC-E dataset.}
\vspace{2mm}
\centering
\begin{tabular}{c|c|ccccl}
\bottomrule
\hline
Dataset                  & Attack    & ASR(\%)$\uparrow$ & RDBLEU$\uparrow$ & RDchrF$\uparrow$ & Sim.$\uparrow$ & Perp.$\downarrow$    \\ \hline
\multirow{3}{*}{VIST-E}  &  Iterative-attack       & 50.37   & 0.49   & 0.23   & 0.96 & 82.95 \\
                         & Text-attack  & 38.46   & 0.34   & 0.12   & 0.95 & 94.94  \\
                         & Image-attack & 22.45   & 0.24   & 0.10    & 1.00 & 79.33  \\ \hline
\multirow{3}{*}{LSMDC-E} &  Iterative-attack       & 47.18   & 0.42   & 0.22   & 0.96 & 176.31 \\
                         & Text-attack  & 45.59   & 0.58   & 0.23   & 0.96 & 177.24 \\
                         & Image-attack & 15.06   & 0.16   & 0.05   & 1.00 & 128.17 \\ \hline
\end{tabular}
\label{tab:tab_ablation_modality}
\end{table*}

\begin{table*}[ht]
\caption{Results of comparison between different level text perturbations in our method against MGCL on VIST-E and LSMDC-E dataset.}
\vspace{2mm}
\centering
\begin{tabular}{c|c|ccccl}
\bottomrule
\hline
Dataset                  & Attack        & ASR(\%)$\uparrow$ & RDBLEU$\uparrow$ & RDchrF$\uparrow$ & Sim.$\uparrow$ & Perp.$\downarrow$ \\ \hline
\multirow{3}{*}{VIST-E}  &  Iterative-attack           & 50.37   & 0.49   & 0.23   & 0.96 & 82.95\\
                         & Character-attack & 48.00   & 0.40   & 0.20   & 0.96 & 88.75 \\
                         & Word-Attack      & 46.32   & 0.44   & 0.22   & 0.97 & 85.54\\ \hline
\multirow{3}{*}{LSMDC-E} & Iterative-attack           & 47.18   & 0.42   & 0.22   & 0.96 & 176.31\\
                         & Character-attack & 45.21   & 0.57   & 0.20   & 0.96 & 180.67 \\
                         & Word-attack      & 46.83   & 0.58   & 0.21   & 0.97 & 186.09\\ \hline
\bottomrule
\end{tabular}
\label{tab:tab_ablation_buggers}
\end{table*}

\subsection{Ablation Study}
In this section, we compare the variants of  Iterative-attack against MGCL on VIST-E and LSMDC-E datasets as shown in Table~\ref{tab:tab_ablation_modality} and Table~\ref{tab:tab_ablation_buggers}, respectively.
From the Table~\ref{tab:tab_ablation_modality} and Table~\ref{tab:tab_ablation_buggers}, we can have the following observations:
\begin{itemize}
    \item [(1)] Perturbing multimodal input iteratively (Iteratively-attack) is consistently stronger than perturbing any single-model input (Text-attack and Image-attack), which demonstrates that the adversarial samples generated by multimodal adversarial attacks are more dangerous than those generated by single-moda adversarial attacks.
    \item [(2)] The performance of Text-attack is better than that of Image-attack on two datasets, however, without the image adversarial attack perturbing the critical visual information, Text-attack performers worse than Iterative-attack, which suggests that due to the complementarity between multimodal data, the information shift caused by single-modal adversarial attack can be corrected by another modality data, thus leading to attack failure.
    \item [(3)] Iterative-attack outperforms Character-attack and Word-attack, which shows that generating substitutes in character-level perturbation and word-replacement via BERT in word-level perturbation together have a greater impact than single-level text perturbation.
\end{itemize}

\begin{table}[ht]
\caption{The results of using different words perturbation number $P$ in the attacking processing.}
\vspace{2mm}
\centering
\begin{tabular}{c|cccc}
\bottomrule
\hline
P & ASR(\%)$\uparrow$                      & Sim. $\uparrow$                       & Perp. $\downarrow$                       & Runtime$\downarrow$ \\ \hline
1 & 54.00                        & 0.98                        & 75.65                        & 59.88   \\ \hline
2 & 66.00                        & 0.96                        & 87.73                        & 75.39   \\ \hline
3 & {\color[HTML]{000000} 71.00} & {\color[HTML]{000000} 0.95} & {\color[HTML]{000000} 93.11} & 77.58   \\ \hline
4 & 80.00                        & 0.95                        & 97.36                        & 100.26  \\ \hline
5 & 82.00                        & 0.93                        & 105.77                       & 107.40  \\ \hline
\bottomrule
\end{tabular}
\centering
\label{tab:num_of_perturbing_words}
\end{table}
\subsection{Visualization}



To further understand Iterative-attack intuitively, we provide the Grad-CAM\footnote{https://github.com/frgfm/torch-cam} \cite{selvaraju2017grad} visualizations for Iterative-attack against MMT on the VIST-E dataset.  The tokens modified by Iterative-attack are written in red in the original story contexts, and the changes in the images are shown by the heat map.

Compared the original image with an adversarial image in Fig. \ref{fig:visualization_MMT_1}, Fig.~\ref{fig:visualization_MMT_2}, we can observe that the word in the adversarial samples has only slightly changed with semantics preservation, but the focus of the visual extractor in the target model on the adversarial images has changed significantly, although these perturbations are imperceptible to humans, which strongly suggests 
the perturbation added to the original image successfully mislead the attention of the target model. Meanwhile, the perturbed percentage for the adversarial context is also low, which can result in more semantic consistency.

To further analyze the proposed method, we added the error analysis. When analyzing instances of failed Iterative-attack attempts, a common observation is that these failures often stem from a fundamental issue: the target model's inability to comprehend the input image effectively. In other words, the ending of the story generated by the model lacks a robust semantic connection with the associated image. 
In Fig.~\ref{fig:visualization_bad_case}, we provide a side-by-side comparison of the original sample and the adversarial sample. Notably, we observe that two words have been altered in the original story context, yet the focus of the visual extractor within the target model remains almost virtually unchanged when examining the adversarial images. Furthermore, when comparing the story ending generated by the model to the ground truth label, it becomes evident that there is a distinct lack of correlation between the story ending and the associated images. This holds true both when the model is provided with clean samples and when it encounters adversarial samples. These findings collectively point to a deficiency in the target model's capacity to effectively comprehend multimodal information, which stands as a pivotal factor constraining the efficacy of the Iterative-attack strategy.


\subsection{Runtime Comparison}
\begin{figure}[ht]
\centering\includegraphics[width=3.5in]{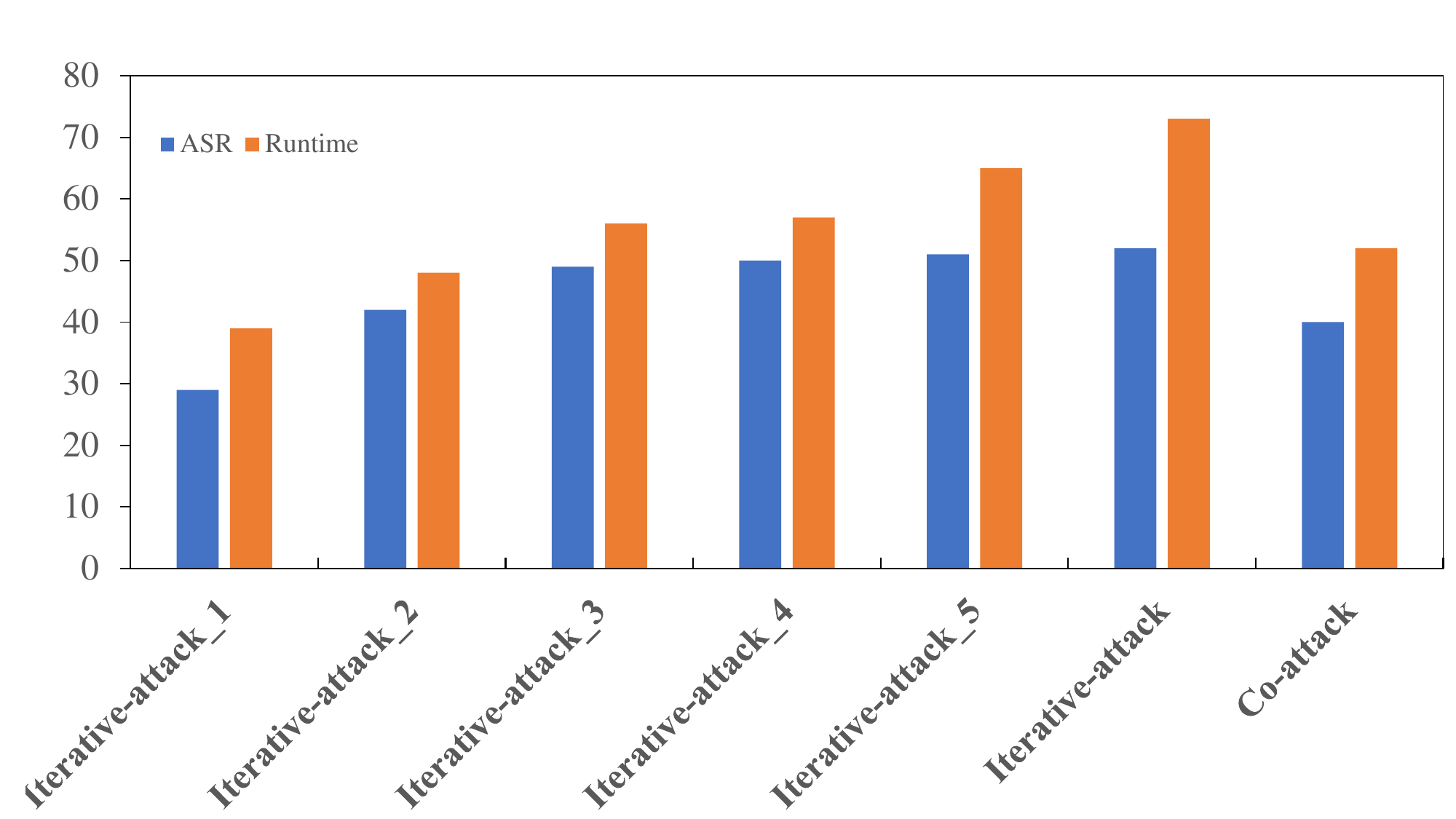}
\caption{The runtime comparison between Iterative-attack with different important word numbers K and Co-attack when attacking MGCL on VIST-E dataset. Iteration-attack\_X means the selected important word number in Iteration-attack is X. ASR means the attack success rate. Runtime is in seconds.}
 \label{fig:cost_time}
\end{figure}
Since kNN and CharSwap are text modality adversarial attack methods, for a fair comparison, we show the average runtime of Iterative-attack with different important word numbers K and Co-attack when attacking on MGCL in the first 100 stories of VIST-E dataset to generate an adversarial sample in Fig.~\ref{fig:cost_time}. We can observe that the runtime of Iterative-attack is slightly higher than Co-attack, yet in the same order of magnitude. What's more, when the runtime is close between Iteration-attack and Co-attack, the ASR score of Iteration-attack is not worse than Co-attack.  

\subsection{Effect on The Number of Perturbed Words in Text}
To verify the effect of using different words perturbation number $P$ in Iterative-attack, we use the first 100 stories and vary the number of perturbing words $P$ from 1 to 5 to attack MGCL on VIST-E dataset, as shown in Table~\ref{tab:num_of_perturbing_words}. We can find that the ASR rises as $P$ increases, but the similarity and fluency of the generated adversarial text are deteriorating, besides, the runtime of attacking is also gradually increasing, which indicates that we need to make a trade-off between the attack success rate and the number of perturbed words in the adversarial attack.

\subsection{Multimodal Adversarial Attack against Multimodal Machine Translation}
To further analyze the effectiveness of Itertaive-attack on multimodal text generation tasks, we test our proposed method with a multimodal machine translation dataset.

\begin{table*}[ht]
\caption{Performance of adversarial attacks against different multimodal machine translation models on Multi30K dataset.}
\vspace{2mm}
\centering
\begin{tabular}{c|c|c|ccccc}
\hline
Dataset                   & Method                       & Attack                    & ASR(\%)$\uparrow$        & RDBLEU$\uparrow$        & RDchrF$\uparrow$        & Sim.$\uparrow$          & Perp.$\downarrow$           \\ \hline
\multirow{8}{*}{Multi30K} & \multirow{4}{*}{MTMMT}       & \textbf{Iterative-attack} & \textbf{44.71} & \textbf{0.40} & \textbf{0.20} & 0.94          & 428.11          \\
                          &                              & Co-attack                 & 20.89          & 0.26          & 0.15          & 0.94          & \textbf{407.38} \\
                          &                              & kNN                       & 19.90          & 0.25          & 0.14          & \textbf{0.96} & 479.16          \\
                          &                              & CharSwap                  & 20.10          & 0.25          & 0.13          & \textbf{0.96} & 425.16          \\ \cline{2-8} 
                          & \multirow{4}{*}{MTMMT${_{back}}$}& \textbf{Iterative-attack} & \textbf{45.40} & \textbf{0.41} & \textbf{0.21} & 0.94          & 442.34          \\
                          &                              & Co-attack                 & 23.50          & 0.29          & 0.16          & 0.94          & \textbf{410.09} \\
                          &                              & kNN                       & 19.30          & 0.22          & 0.12          & \textbf{0.97} & 490.53          \\
                          &                              & CharSwap                  & 19.30          & 0.22          & 0.12          & 0.96          & 490.99          \\ \hline
\end{tabular}
\centering
\label{tab:attack_machine_translation}
\end{table*}

\begin{table*}[ht]
\caption{Results of ablation experiments for the multimodal machine translation models on Multi30K dataset.}
\vspace{2mm}
\centering
\begin{tabular}{c|c|c|ccccc}
\hline
Dataset                    & Model                        & Attack           & ASR(\%) & RDBLEU & RDchrF & Sim. & Perp.  \\ \hline
\multirow{10}{*}{Multi30K} & \multirow{5}{*}{MTMMT}       & Iterative-attack & 44.71   & 0.40   & 0.20   & 0.94 & 428.11 \\
                           &                              & Text-attack      & 43.56   & 0.39   & 0.20   & 0.93 & 403.27 \\
                           &                              & Image-attack     & 0.99    & 0.01   & 0.01   & 1.00 & 180.36 \\
                           &                              & Character-attack & 37.19& 0.37   & 0.19   & 0.93 & 433.62 \\
                           &                              & Word-attack      & 35.37   & 0.33   & 0.17   & 0.95 & 387.55 \\ \cline{2-8} 
                           & \multirow{5}{*}{MTMMT${_{back}}$} & Iterative-attack & 45.40   & 0.41   & 0.21   & 0.94 & 442.34 \\
                           &                              & Text-attack      & 44.80   & 0.39   & 0.20   & 0.94 & 418.65 \\
                           &                              & Image-attack     & 1.48    & 0.02& 0.02   & 1.00 & 180.36 \\
                           &                              & Character-attack & 37.90   & 0.36   & 0.19   & 0.93 & 470.85 \\
                           &                              & Word-attack      & 39.02& 0.36   & 0.19   & 0.95 & 390.34 \\ \hline
\end{tabular}
\centering
\label{tab:attack_machine_translation_ablation}
\end{table*}

\begin{table}[ht]
\caption{Performance of the multimodal machine translation models on Multi30K dataset.}
\vspace{2mm}
\centering
\begin{tabular}{c|c|ccccc}
\hline
Dataset                   & Method        & B1   & B2    & B3    & B4    & Meteor \\ \hline
\multirow{4}{*}{Multi30K} & MTMMT\#       & -    & -     & -     & 38.70 & 55.70  \\
                          & MTMMT*         & 67.6 & 54.91 & 45.81 & 38.65 & 55.06  \\
                          & MTMMT${_{back}}$\# & -    & -     & -     & 39.50 & 56.90  \\
                          & MTMMT${_{back}}$*   & 68.2 & 55.26 & 47.15 & 39.48 & 56.88  \\ \hline
\end{tabular}
\centering
\begin{threeparttable}
* indicates the actual results of running. \# indicates the results in the paper.
\end{threeparttable}
\label{tab:real_machine_translation}
\end{table}

\subsubsection{Multimodal Machine Translation Models}

MTMMT~\cite{yao2020multimodal} is a multimodal machine translation model that uses multimodal self-attention  in Transformer to avoid encoding irrelevant information in images. To further investigate the performance of MTMMT on more data, the authors also train the MTMMT with the additional training data where the authors use a back-translation model~\cite{sennrich2016improving} to translate 145k monolingual German description in M30kC into English, and the model refers to MTMMT${_{back}}$. The reproducible results are reported in Table~\ref{tab:real_machine_translation}
\subsubsection{Quantitative Results}

Table~\ref{tab:attack_machine_translation} and Table~\ref{tab:attack_machine_translation_ablation} presents the results of attacks on two multimodal machine translation models within the Multi30K dataset. The following observations are noteworthy:
\begin{itemize}
    \item [(1)] Comparative Efficacy of Iterative-Attack: Our Iterative-attack method outperforms other baseline approaches in terms of attack success rate and the extent to which translation quality is reduced. This is evident from the close semantic similarity and perplexity scores. Such results underscore the superior performance of our proposed attack method in both the IgSEG task and the multimodal machine translation task.
    \item [(2)] Compared the performance of multimodal attack methods (Iterative-attack and Co-attack) with single-modal attack methods (kNN and CharSwap), it is obvious that multimodal attack methods outperforms the single-modal attack methods, which proves our observation that for the multimodal text generation models, the single-modal perturbation tends to fail, due to the complementary information between text data and image
    \item [(3)] Effectiveness of Multimodal Attacks in Text Generation: In multimodal text generation tasks, the integration of textual and visual data plays a significant role in producing coherent texts. Our findings suggest that combined attacks on both texts and images are more effective than attacks on a single modality.
\end{itemize}
These insights reflect a comprehensive understanding of the vulnerabilities and characteristics of multimodal machine translation models, providing a foundation for further research in this domain.

\subsubsection{Ablation Study}

In this section, we compare the variants of  Iterative-attack against MTMMT and MTMMT${_{back}}$ on Multi30K dataset as shown in Table~\ref{tab:attack_machine_translation_ablation}. 
From the Table~\ref{tab:attack_machine_translation_ablation}, we can have the following observations:
\begin{itemize}
    \item [(1)] Perturbing multimodal input iteratively (Iteratively-attack) is consistently stronger than perturbing any single-model input (Text-attack and Image-attack), which demonstrates that the adversarial samples generated by multimodal adversarial attacks are more dangerous than those generated by single-moda adversarial attacks in multimodal machine translation.
    \item [(2)] The Iterative-attack, which consists of both character-level and word-level perturbations, significantly outperforms either Character-attack or Word-attack alone. This outcome highlights the amplified impact achieved by integrating character-level substitutes and BERT-based word-replacement strategies, as opposed to focusing on a single level of text perturbation in multimodal text generation tasks. 
    \item [(3)] In multimodal machine translation tasks, Text-attack performs better than Image-attack. This is attributed to the distinct role of images in this domain, primarily serving to enhance contextual understanding and translation accuracy. Unlike in IgSEG tasks, image perturbations in multimodal machine translation frequently fail to significantly influence the output of the target models, reflecting the limited impact of visual modifications on the translation process. 
\end{itemize}

\subsection{Discussion}
To further understand why multimodal adversarial examples outperform single-modal adversarial samples, what makes the proposed method fail, and the practical implications of our research, we discuss some key concepts of adversarial attacks in this paper that encompass both empirical and theoretical elements. In detail:
\begin{itemize}
    \item For a multimodal model,  perturbing bi-modal inputs is stronger than perturbing any single-modal input~\cite{zhang2022towards}. In the experiment of adversarial attack on the multimodal text generation models, the single-modal perturbation tends to fail, due to the complementary information between text data and image. The multimodal adversarial attack can find the most vulnerable multimodal adversarial patches to avoid the dilemma that the information shift caused by a single-modal adversarial attack may be corrected by another modality’s information.
    \item A robust correlation exists between the effectiveness of the Iterative-attack and the target model's capacity to interpret multimodal information holistically. In our observations, we have a conclusion that if the ending of a story produced by the target model exhibits no meaningful connection to the accompanying image, perturbations applied to the image do not exert a discernible influence on the model's output, which suggests that the target model's ability to understand multimodal information is important for the success of  Iterative-attack.
    \item Multimodal text generation represents a pivotal challenge in artificial intelligence, requiring the integration and interpretation of diverse information sources to produce coherent textual outputs. Key applications of multimodal text generation include multimodal machine translation~\cite{su2019unsupervised, liu2021gumbel}, multimodal dialogue response generation~\cite{sun2021multimodal, wang2021modeling}, multimodal question answering~\cite{singh2021mimoqa}, multimodal MemexQA~\cite{47871}, and image-guided story ending generation~\cite{huang2021igseg, xue2022mmt}. These diverse applications underscore the versatility and complexity of multimodal text generation, highlighting its significance in advancing the field of artificial intelligence. 
    Our study contributes significantly to the unexplored domain of adversarial robustness in multimodal text generation systems. By investigating the adversarial robustness of multimodal text generation models to multimodal adversarial attacks, this research accomplishes two primary objectives: firstly, it elucidates the internal mechanisms of these complex models; secondly, it aids in the development of more robust and reliable multimodal systems. 
\end{itemize}

\subsection{Insights and important conclusions }
we present critical observations and conclusions drawn from adversarial attacks against image-guided story ending generation (IgSEG) and multimodal machine translation tasks. The key conclusions are as follows: 
\begin{itemize}
    \item \textbf{Effectiveness of Iterative-Attack Methodology}: Our analysis demonstrates that Iterative-attack consistently outperforms other attack methods (such as Co-attack, kNN, and CharSwap) in achieving a higher attack success rate (ASR) across different datasets and models. This suggests that the Iterative-attack is more effective at finding and exploiting vulnerabilities in IgSEG models and multimodal machine translation models. 
    \item \textbf{Impact of Attack Multi-modality}: In the context of multimodal text generation, the effectiveness of a single-modal attack is likely to be affected by the complementary information in texts and images. Our proposed Iterative-attack, which iteratively attacks text and image modalities, consistently shows higher ASR compared to image adversarial attacks and text adversarial attacks alone. Additionally, adversarial attacks on classification tasks seek to induce incorrect labels, exploiting vulnerabilities around the decision boundary of the classifier. In contrast, adversarial attacks on multimodal text generation tasks aim to disrupt the generation process, leading to outputs that may be grammatically correct but are contextually inappropriate or misleading. The latter is more difficult due to the complexity and variability of language generation. 
    \item \textbf{Vulnerability of Multimodal Text Generation Models}: The varying performance of different IgSEG models (such as Seq2Seq, Transformer, MGCL, MMT) and multimodal machine translation models (like MTMMT) to various attack strategies suggests a fundamental vulnerability. Current methodologies for integrating and processing multimodal inputs (text and images) for text generation are not sufficiently robust against adversarial attacks. This indicates a need for more advanced models capable of effectively fusing multimodal information, potentially enhancing resilience to such attacks. 
\end{itemize}

\section{Conclusion}
\label{sec:conclusion}
In this paper, we first propose an iterative multimodal adversarial attack against IgSEG models in multimodal text generation tasks. Compared with the single-modal adversarial attack methods, our method 
fuses the image modality attack into the text modality attack to iteratively find the most vulnerable multimodal information patch. The generated multimodal adversarial samples can avoid the dilemma that the information shift caused by a single-modal adversarial attack may be corrected by another modality’s information.
Our experimental results show that Iterative-attack is highly effective against multimodal text generation tasks.

Evaluating the adversarial robustness within the realm of multimodal text generation assumes paramount importance when considering the pragmatic deployment of multimodal models. By employing iterative attacks on the target model, we can systematically identify and comprehend the inherent vulnerabilities within the multimodal architecture. This, in turn, motivates us to institute comprehensive measures aimed at fortifying its robustness. 
In the future, we plan to establish a comprehensive and rigorous benchmark to evaluate adversarial robustness on multimodal text generation for different applications and investigate methods for defending against multimodal adversarial attacks.
Besides, we plan to establish a comprehensive and rigorous benchmark to evaluate adversarial robustness on multimodal text generation for different applications and investigate a method for defending against multimodal adversarial attacks.

We hope that this study will draw attention to the adversarial robustness of multimodal text generation tasks.

\bibliographystyle{IEEEtran}
\bibliography{egbib.bib}

\begin{IEEEbiography}[{\includegraphics[width=1in,height=1.25in,clip,keepaspectratio]{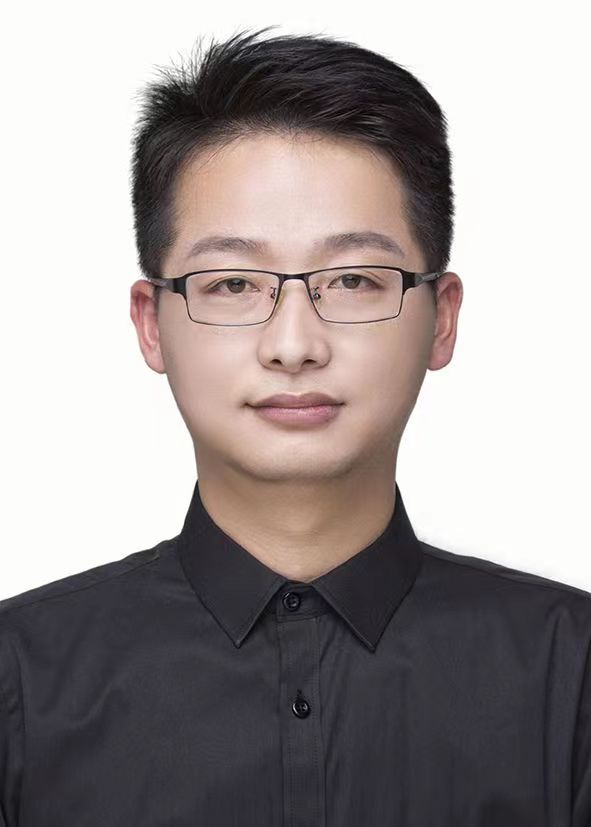}}]{Youze Wang}
received a B.S. and master's degree from the School of Computer Science and Information Engineering at Hefei University of Technology, Hefei, China, where he is currently working toward his Ph.D. degree. His research interests include multimodal computing and multimodal adversarial robustness in machine learning. 
\end{IEEEbiography}

\begin{IEEEbiography}[{\includegraphics[width=1in,height=1.25in, clip,keepaspectratio]{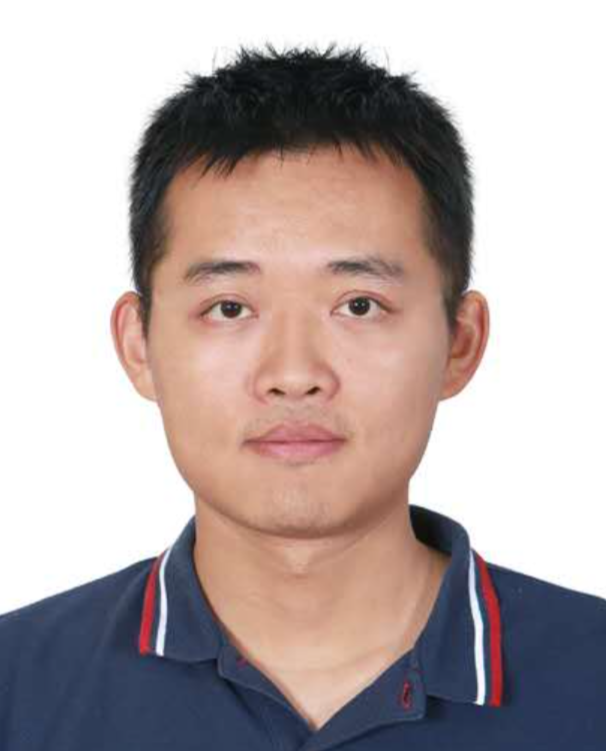}}]{Wenbo Hu} is an associate professor in Hefei University of Technology. He received a Ph.D. degree from Tsinghua University in 2018. His research interests lie in machine learning, especially probabilistic machine learning and uncertainty, generative AI, and AI security. He has published more than 20 peer-reviewed papers in prestigious conferences and journals, including NeurIPS, KDD, IJCAI, etc.
\end{IEEEbiography}

\begin{IEEEbiography}[{\includegraphics[width=1in,height=1.25in,clip,keepaspectratio]{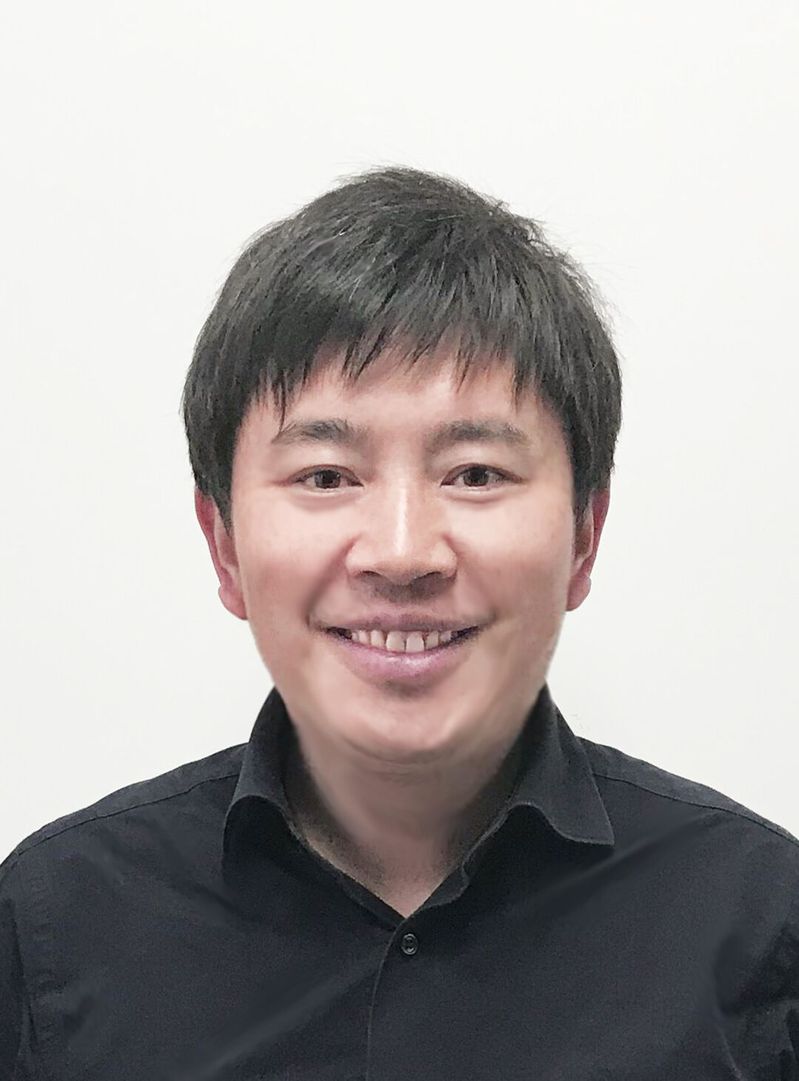}}]{Richang Hong (Member, IEEE)}
received a Ph.D. degree from the University of Science and Technology of China, Hefei, China, in 2008. He was a Research Fellow of the School of Computing at the National University of Singapore, from 2008 to 2010. He is currently a Professor at the Hefei University of Technology, Hefei. He is also with the Key Laboratory of Knowledge Engineering with Big Data (Hefei University of Technology), Ministry of Education. He has coauthored over 100 publications in the areas of his research interests, which include multimedia content analysis and social media. He is a member of the ACM and the Executive Committee Member of the ACM SIGMM China Chapter. He was a recipient of the Best Paper Award from the ACM Multimedia 2010, the Best Paper Award from the ACM ICMR 2015, and the Honorable Mention of the IEEE Transactions on Multimedia Best Paper Award. He has served as the Technical Program Chair of the MMM 2016, ICIMCS 2017, and PCM 2018. Currently, he is an Associate Editor of IEEE Transactions on Big Data, IEEE Transactions on Computational Social Systems, ACM Transactions on Multimedia Computing Communications and Applications, Information Sciences (Elsevier), Neural Processing Letter (Springer) and Signal Processing (Elsevier).
\end{IEEEbiography}

 




\vfill

\end{document}